\documentclass[conference]{IEEEtran}

\IEEEoverridecommandlockouts
\usepackage{cite}
\usepackage{amsmath,amssymb,amsfonts}
\usepackage{textcomp}
\def\BibTeX{{\rm B\kern-.05em{\sc i\kern-.025em b}\kern-.08em
    T\kern-.1667em\lower.7ex\hbox{E}\kern-.125emX}}

\usepackage{multirow}
\usepackage{adjustbox}
\usepackage{subcaption}

\usepackage{graphicx}
\usepackage[table,xcdraw]{xcolor}
\usepackage{algpseudocode}
\usepackage{algorithm}
\usepackage{hyperref}
\hypersetup{colorlinks=true,
            urlcolor=[RGB]{30, 144, 80},
            linkcolor=[RGB]{211, 64, 54},
            citecolor=[RGB]{5, 125, 238}}
\newcommand{\etal}{\textit{et al}. }
\newcommand{\ie}{\textit{i}.\textit{e}., }
\newcommand{\eg}{\textit{e}.\textit{g}. }

\begin{document}

\title{GraphGDP: Generative Diffusion Processes for Permutation Invariant Graph Generation
}


\author{
\IEEEauthorblockN{Han Huang$^1$, Leilei Sun$^{1*}$, Bowen Du$^1$, Yanjie Fu$^2$, Weifeng Lv$^1$}
\IEEEauthorblockA{
$^1$\textit{SKLSDE,
Beihang University, Beijing, China}\\
$^2$\textit{University of Central Florida, Florida, USA}\\
Email: \{h-huang, leileisun, dubowen, lwf\}@buaa.edu.cn, yanjie.fu@ucf.edu}
\thanks{$^*$ Corresponding author.}
\thanks{$\bullet$ Code is available at \url{https://github.com/GRAPH-0/GraphGDP}.}
}%

\maketitle

\begin{abstract}
Graph generative models have broad applications in biology, chemistry and social science. 
However, modelling and understanding the generative process of graphs is challenging due to the discrete and high-dimensional nature of graphs, as well as permutation invariance to node orderings in underlying graph distributions.
Current leading autoregressive models fail to capture the permutation invariance nature of graphs for the reliance on generation ordering and have high time complexity.
Here, we propose a continuous-time generative diffusion process for permutation invariant graph generation to mitigate these issues.
Specifically, we first construct a forward diffusion process defined by a stochastic differential equation (SDE), which smoothly converts graphs within the complex distribution to random graphs that follow a known edge probability.
Solving the corresponding reverse-time SDE, graphs can be generated from newly sampled random graphs.
To facilitate the reverse-time SDE, we newly design a position-enhanced graph score network,  capturing the evolving structure and position information from perturbed graphs for permutation equivariant score estimation.
Under the evaluation of comprehensive metrics, our proposed generative diffusion process achieves competitive performance in graph distribution learning.
Experimental results also show that GraphGDP can generate high-quality graphs in only 24 function evaluations, much faster than previous autoregressive models.
\end{abstract}


\section{Introduction} \label{section1}

\begin{figure*}[t]
    \centering
    \adjustimage{width=0.97\textwidth, center}{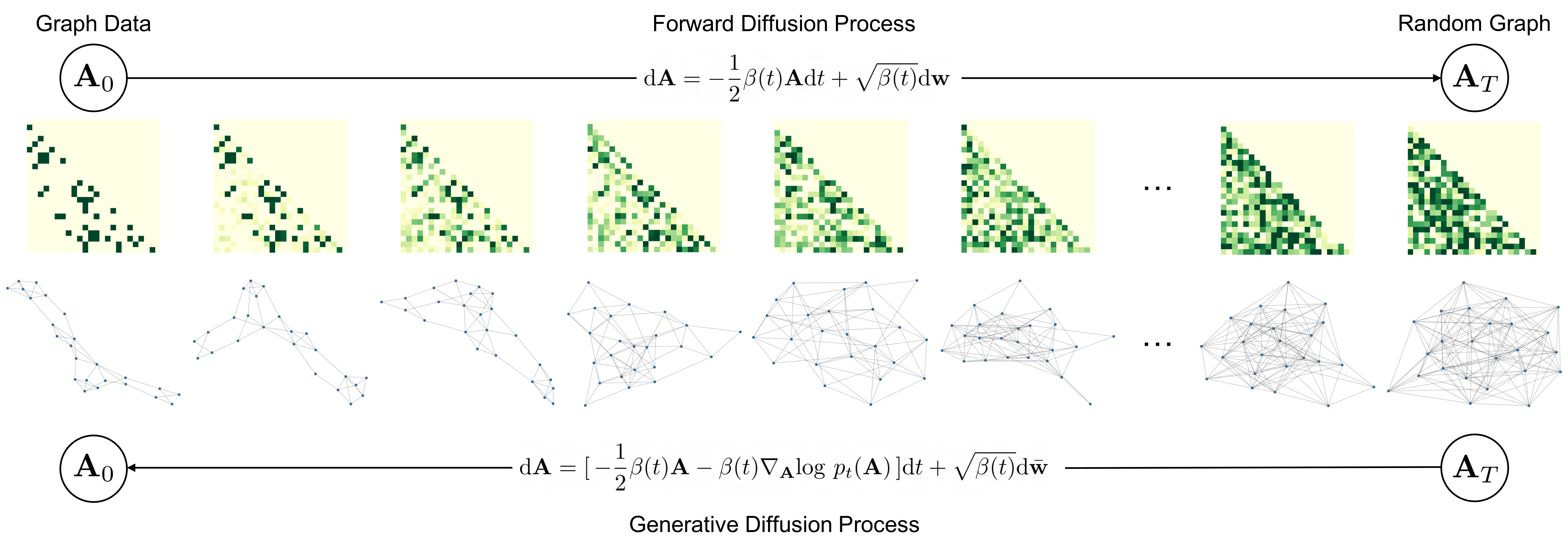}
    \caption{Perturbing graph data to random graphs with an underlying simple edge distribution can be achieved by a continuous-time forward diffusion process described by an SDE.
    We can convert this SDE with the score of the data distribution $\nabla_{\mathbf{A}}\mathrm{log}\ p_{t}(\mathbf{A})$ at each time $t$ .
    Thus, a graph sample can be transformed from a newly sampled random graph via the generative diffusion process.
    The lower triangle part of adjacency matrices perturbed from the same graph at different times are shown in the first row, and the corresponding discrete graphs are below them. 
    }
    \label{fig:pipeline}
\end{figure*}


Graph data is a ubiquitous and highly applicable type of high-dimensional data in data mining.
Modelling and understanding the generative process of graphs has been studied for a long time in network science \cite{newman2018networks} and continues as an active research topic incorporating deep learning \cite{guo2020systematic}.
Graph generative models aim to capture the underlying distributions over a particular family of graphs and generate diverse novel graphs with high fidelity, which serves as the foundation for wide applications, such as de novo drug discovery \cite{li2018learning, zang2020moflow, Shigraphaf20, Luographdf21}, computation graph creation for network architecture design \cite{XieERWNN19}, semantic parsing in natural language \cite{chen2018sequence}, and analysis in network science \cite{watts1998collective, albert2002statistical, leskovec2010kronecker}.


Learning the distribution of discrete and combinatorial graph structures is a challenging task, which is also a necessary and fundamental step for further jointly modelling attributes \cite{zang2020moflow, Shigraphaf20, Luographdf21} and labels \cite{goyal2020graphgen} in semantic abundant graphs.
Traditional methods for graph generation date back to random graph models \cite{erdHos1960evolution, watts1998collective, albert2002statistical}, which rely on hand-crafted stochastic generation processes and capture limited graph statistic properties.
Recent deep graph generative models utilize the capacity of neural networks to learn graph structure distribution effectively.
The prominent paradigms include variational autoencoder (VAE) based models \cite{kipf2016variational, simonovsky2018graphvae, liu2018constrained}, generative adversarial network (GAN) based models \cite{bojchevski2018netgan, de2018molgan}, flow-based models \cite{ nips/LiuKBKS19, zang2020moflow, Shigraphaf20, Luographdf21}, and autoregressive models \cite{li2018learning, you2018graphrnn, nips/LiaoLSWHDUZ19, dai2020scalable, icml/ChenHHRL21}.   
Among them, autoregressive models achieve the most impressive generation quality on discrete graph structures.
However, they rely on node generation orderings with high time complexity and fail to capture the important permutation invariant properties of graphs.
The desired likelihood-based graph generative models should estimate invariant likelihood to all possible equivalent adjacency matrices of the same graph.
To reach this goal, Niu \etal \cite{niu2020permutation} creatively integrate score-based generative models \cite{song2019generative} with graph neural networks to implicitly represent permutation-invariant distributions, but still suffer from generation quality and sampling speed.  



In this paper, we propose continuous-time generative diffusion processes for permutation invariant graph generation (GraphGDP), which exhibit high graph generation quality and efficient sampling potential. 
Analogous to the diffusion processes in a non-equilibrium thermodynamic system in which particles move stochastically under the influence of a heat bath and spread out over the entire space in equilibrium 
\cite{de2013non, sohl2015deep, ho2020denoising},
we perturb the edges in graphs with a sequence of noise (a.k.a, forward diffusion processes), and generate graphs by learning to reverse this process from noise to data (a.k.a, generative diffusion processes).
It is non-trivial to adapt current methods to graph data effectively due to the permutation invariance constraint and the necessity to account both discrete local motifs and overall topological properties of graphs.


Inspired by the seminal work \cite{song2021score} with superior image generation quality, which connects diffusion-based models and score matching by defining the stochastic differential equation (SDE) describing continuous-time perturbing processes and the reverse-time SDE for generation,
we define a forward diffusion process described by the specialized SDE with a closed-form expression over the real-valued adjacency matrices.
We make an important observation that such a forward diffusion process implicitly defines a corresponding conversion process on the discrete distribution of graphs through a simple quantization on perturbed adjacency matrices.
During the diffusion process, the edge existing probability in the Bernoulli distribution evolves stochastically over time, and the signal-to-noise ratio of the graph continues to decrease.
In the final equilibrium state, the graph distributions are converted to Erd{\H{o}}s-R{\'e}nyi random graphs \cite{erdHos1960evolution} with an edge probability of $0.5$.
The correlation between continuous-valued adjacency matrices and discrete graphs provides not only an intuitive understanding of the diffusion process on graphs, but also extra topology information for denoising in the reverse process.

Our ultimate goal is to recover the complex graph distribution from the prior noise distribution through generative diffusion processes.   
The success of reverse-time SDEs is heavily dependent on the neural networks (named score networks) to estimate scores (a.k.a., the gradient field of the perturbed data distribution).
To capture the evolving graph topology, we emphasize the use of discrete graph states quantized from perturbed continuous-valued adjacency matrices.
Intuitively, the evolving node or edge position information is critical in graph structure denoising.
We propose an effective position-enhanced graph score network (PGSN) for permutation equivariant edge score estimation, which extracts structure and position features from quantized graphs and combines them with dense continuous adjacency matrices.
With the optimized PGSN, we utilize numerical solvers for reverse-time SDEs and obtain final graph samples along approximate trajectories.
For the graph generation quality evaluation, we apply metrics based on recent works \cite{o2021evaluation, thompson2022Metric} to benchmark graph generative models on widely used datasets, and we find that our model achieves better or comparable performance to strong autoregressive models.
Using the probability flow ordinary differential equation (ODE) sharing marginal probability densities with the SDE, we also show that our model can achieve efficient graph sampling with considerable quality in only $24$ function evaluations using ODE solvers.
The framework is summarized in Fig. \ref{fig:pipeline}. 

Our main contributions are summarized as:
\begin{itemize}
\item We propose a novel continuous-time generative diffusion process for permutation invariant graph generation, which incorporates observations of continuous-valued adjacency matrices and corresponding discrete graphs for the interconversion between graph samples and random graphs.
\item We design a position-enhanced graph score network that accurately estimates scores on perturbed graphs by leveraging the perturbed adjacency matrices as well as the structure and position information of discrete graphs.
\item With the carefully designed evaluation setting, our proposed model achieves better or comparable sampling quality to autoregressive models in graph generation and displays strong potential for efficient sampling.
\end{itemize}

\section{Preliminaries} \label{section2}

\subsection{Continuous-time Generative Diffusion Processes} \label{subsec:CGDP} 
For a datapoint $\mathbf{x} \in \mathbb{R}^{d}$ , consider a forward diffusion process $\mathbf{x}_t$ defined by an It\^{o} SDE:
\begin{equation}
    \mathrm{d} \mathbf{x} = \mathbf{f}(\mathbf{x}, t)\mathrm{d}t + \mathbf{G}(\mathbf{x}, t)\mathrm{d} \mathbf{w} \ , 
\label{eq:forward_sde}
\end{equation}
with an indexed continuous time variable $t \in [0, T]$, a standard Wiener process $\mathbf{w}$ (a.k.a., Brownian motion) ,  a drift coefficient $\mathbf{f}(\cdot, t): \mathbb{R}^d \to \mathbb{R}^d$ and a diffusion coefficient $\mathbf{G}(\cdot, t) : \mathbb{R}^d \to \mathbb{R}^{d \times d}$.
According to \cite{anderson1982reverse, haussmann1986time, song2021score}, running backwards in time from $T$ to $0$ (\ie with negative $dt$), a corresponding reverse-time diffusion process inverting the above forward diffusion process can derived as:
\begin{equation}
\begin{aligned}
\mathrm{d}{\mathbf{x}} & = \{ 
\mathbf{f}(\mathbf{x}, t) - 
\nabla \cdot [ \, \mathbf{G}(\mathbf{x}, t) \mathbf{G}(\mathbf{x}, t)^{\top} \,] \\
& - \mathbf{G}(\mathbf{x}, t)\mathbf{G}(\mathbf{x}, t)^{\top} \nabla_{\mathbf{x}}\mathrm{log}p_t(\mathbf{x})
\}\mathrm{d}t
+ \mathbf{G}(\mathbf{x}, t)\mathrm{d} \bar{\mathbf{w}} \ , 
\end{aligned}
\label{eq:reverse_sde}
\end{equation}
where $\nabla_{\mathbf{x}}\mathrm{log}p_t(\mathbf{x})$ is the score function of the marginal distribution over data $\mathbf{x}$ at time $t$ and $\bar{\mathbf{w}}$ is  a reverse-time standard Wiener process. 

Song \etal \cite{song2021score} show that the reverse-time diffusion process can be converted into a generative model known as the continuous-time generative diffusion process.
The SDEs commonly used for diffusion take the simple form for drift and diffusion coefficients with $\mathbf{f}(\mathbf{x}, t)=f(t)\mathbf{x}_t$ and $\mathbf{G}(\mathbf{x}, t)=g(t)\mathbf{I}$.
With some specific designs for $f(t)$ and $g(t)$, the marginal and equilibrium density of the SDE approximates a Normal distribution at time $T$,  \ie $\mathbf{x}_T \sim \mathcal{N}(\mathbf{x}_T; \mathbf{0}, \mathbf{I})$.  \footnote{We only use the "variance-preserving" SDEs in this paper, while there are other SDEs with different Gaussian distributions in equilibrium.}
Initializing $\mathbf{x}_0$ with a sample from the complex data distribution, the state $\mathbf{x}_t$ gradually approaches equilibrium via Eq. (\ref{eq:forward_sde}).
After training a time-dependent score network $\mathbf{s_\theta}(\mathbf{x}_t, t)$, \ie the parametric score function, for estimating the score $\nabla_{\mathbf{x}_t}\mathrm{log}\ p_t(\mathbf{x}_t)$, we can synthesize data via the reverse-time SDE in Eq. (\ref{eq:reverse_sde}).
The denoising score matching \cite{vincent2011connection} objective is modified for score estimation training as:
\begin{equation}
    \min_{\mathbf{\theta}}\mathbb{E}_{t}\{
    \lambda(t) \mathbb{E}_{\mathbf{x}_0} \mathbb{E}_{\mathbf{x}_t | \mathbf{x}_0}
    [\,||\mathbf{s_\theta}(\mathbf{x}_t, t) - \nabla_{\mathbf{x}}\mathrm{log}\ p_{0t}(\mathbf{x}_t|\mathbf{x}_0)
    ||_2^2\,]
    \} \ ,
 \label{eq:loss}
\end{equation}
where $\lambda(t)$ is a given positive weighting function.
When $\mathbf{f}(\cdot, t)$ and $\mathbf{G}(\cdot, t)$ are affine, the transition kernel $p_{0t}(\mathbf{x}_t | \mathbf{x}_0)$ keeps a tractable Gaussian distribution \cite{sarkka2019applied}, which helps compute the score target and perturbed data efficiently.

\subsection{Permutation Invariant Generation} 

The goal of graph generative models is to capture the underlying distribution $p(G)$ over graph instances and then sample a new graph from the learned distribution.
A graph $G$ with $n$ nodes is defined as $G = (V, E)$, where $V, E$ correspond to the node set and the edge set.
In this paper, we only consider the undirected graphs without self-loops and multi-edges. 
Given a node ordering $\pi = (\pi_{1}, ..., \pi_{n})$, the graph $G$ is determined by its adjacency matrix $\mathbf{A}^{\pi} \in \mathbb{R}^{n \times n}$.
Therefore, the graph distribution can also be represented by marginalizing over all adjacency matrices. 
We omit the $\pi$ in $\mathbf{A}^{\pi}$ if not emphasizing the node ordering.
Since the graph distribution is inherently invariant to any node permutation, 
graph generative models are expected to estimate the same likelihood to equivalent adjacency matrices.
Niu \etal \cite{niu2020permutation} make a theoretical analysis that if the edge output of the score network is permutation equivariant, then the gradient of log-likelihood estimation $\mathbf{s_\theta}(\mathbf
{A})$ is permutation equivariant and the implicitly defined log-likelihood function $\mathrm{log}\ p_{\mathbf{\theta}} (\mathbf{A})$ is permutation invariant.

\section{Model} \label{section3}

Our approach aims to apply continuous-time generative diffusion processes to graphs, capturing the desirable permutation invariant graph distribution.
Below, we first describe the forward diffusion process that perturbs graph instances towards random graphs in Section \ref{subsec:forward}.
Then we illustrate our specialized position-enhanced graph score network for estimating graph scores (\ie denoising perturbed graphs) in Section \ref{subsec:pgsn}.
The detail of graph generative diffusion processes is provided in Section \ref{subsec:graphgdp}.  

\begin{figure}[t]
    \centering
    \adjustimage{width=\columnwidth, center}{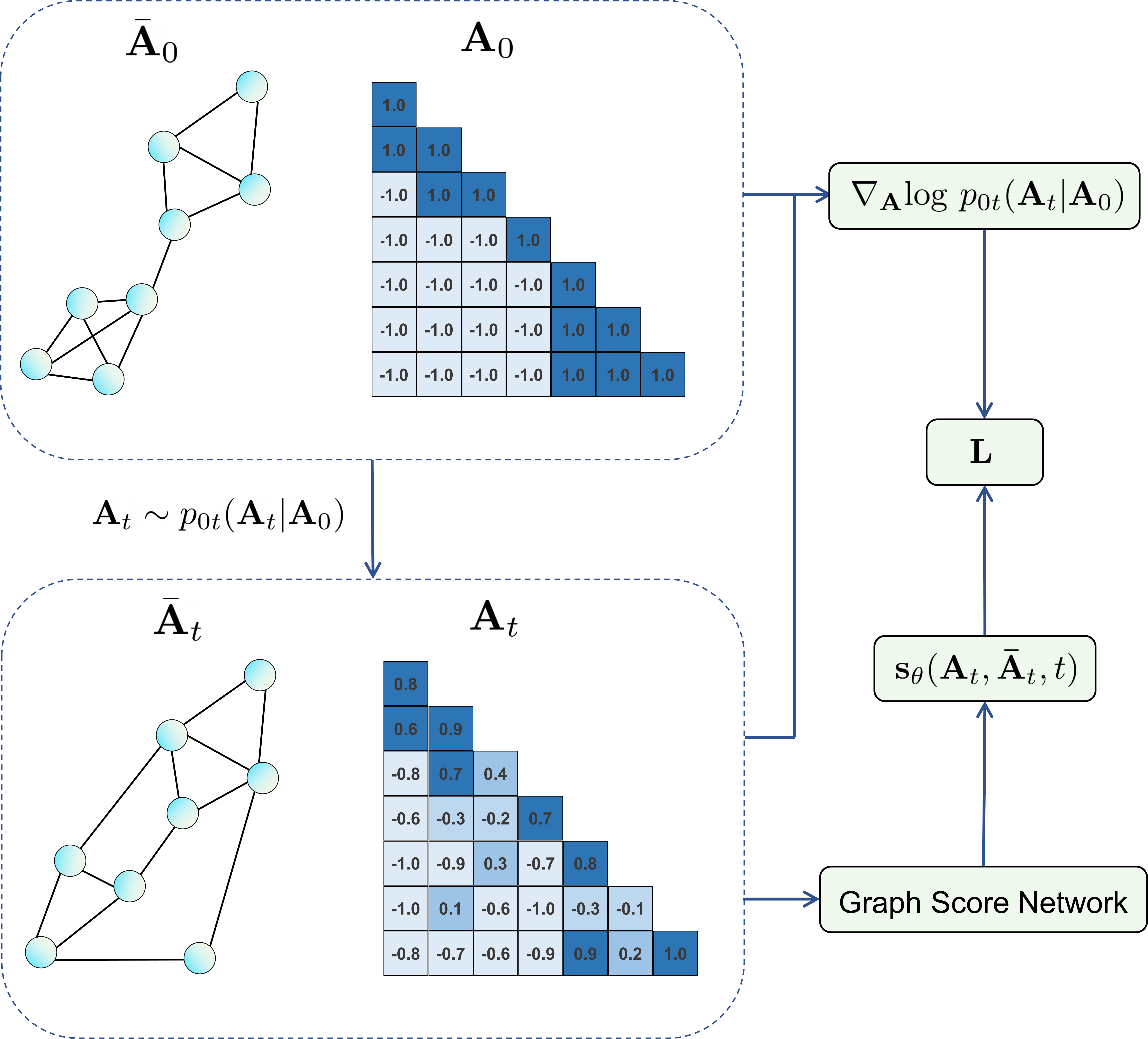}
    \caption{The illustration of the objective computation for the proposed graph score network.
    Uniformly sample $t \in (0,1]$, compute perturbed graphs and the gradient of log probability density at time $t$, and then train graph score networks to estimate scores.}
    \label{fig:train}
\end{figure}

\subsection{Perturbing Graphs Towards Random Graphs} \label{subsec:forward}
Our goal is to construct a diffusion process that perturbs graphs towards the equilibrium state like Eq. (\ref{eq:forward_sde}).
Intuitively, a straightforward but effective method is to add noise into each element in adjacent matrices considering the discrete elements as continuous variables.    
For a graph diffusion process $\mathbf{A}_t$ indexed by a continuous time variable $t \in [0, T]$, we denote $\mathbf{A}_0 \sim p_0$ as an adjacency matrix drawn from the original data distribution $p_0$, and $\mathbf{A}_T \sim p_T$
as the final perturbed adjacency matrix drawn from the prior distribution $p_T$.

The choice of the drift coefficient function $f(t)$ and diffusion coefficient function $g(t)$ is indispensable for the success of diffusion process as shown in Section \ref{subsec:CGDP}.
On the one hand, it should ensure that the corresponding SDE can diffuse the data from a complex high-dimensional distribution into a tractable prior distribution with a low signal-to-noise ratio that is convenient for sampling.
On the other hand, it is assumed to be designed as affine functions that allow us to efficiently sample the state $\mathbf{A}_t$ at any time $t$ in the forward diffusion process without model parameters.
We apply the variance-preserving SDE introduced by \cite{song2021score} for graph data, whose discrete-time form is a denoising diffusion probabilistic model \cite{ho2020denoising}. 
Considering the linear scaling function 
$\beta(t) = \bar{\beta}_{min} + t(\bar{\beta}_{max} - \bar{\beta}_{min})$
for $t \in [\,0, 1\,]$, we derive the SDE from Eq. (\ref{eq:forward_sde}) as:
\begin{equation}
    \mathrm{d}\mathbf{A} = - \frac{1}{2} \beta(t) \mathbf{A} \mathrm{d}t + \sqrt{\beta(t)} \mathrm{d}\mathbf{w} \ .
\label{eq:graphsde}
\end{equation}
In detail, the values of $\mathbf{A}$ are linearly scaled to $[-1, 1]$. 
At time $t$, $\beta(t)$ actually control the scale of adding noise. $\beta(t)$ is shared for all elements in $\mathbf{A}$, while the noise term from $\mathbf{w}$ is independently sampled for each element.
As graphs are undirected here, we manipulate the lower triangular part of adjacency matrices and make them symmetric afterwards.
According to Eq. (\ref{eq:graphsde}), with affine drift coefficients, the perturbation kernels $p_{0t}(\mathbf{A}_t | \mathbf{A}_0)$ are Gaussian distributions as \cite{sarkka2019applied}:
\begin{equation}
 p_{0t}(\mathbf{A}_t | \mathbf{A}_0) = \mathcal{N}(\mathbf{A}_t; 
 \mathbf{A}_0e^{-\frac{1}{2}\int_0^t \beta(s)\mathrm{d}s}, 
 \mathbf{I} - \mathbf{I}e^{-\int_0^t \beta(s)\mathrm{d}s}
 ) \ .
\end{equation}
With these perturbation kernels, we can perturb the graphs and compute the score target efficiently at any time without running the forward diffusion process step by step.

The forward diffusion process on graphs also implicitly defines a corresponding transformation on the discrete distribution, although it manipulates the continuous distribution.
A simple quantization of the sampled $\mathbf{A}_t$ (\ie with the discretization threshold of $0.5$ for the edge values) converts it to a graph denoted by $\mathbf{\bar{A}}_t$ that is sampled from the $\{0, 1\}^{n \times n}$ Bernoulli distribution with gradually erased original signals.       
The final equilibrium state $\mathbf{\bar{A}}_T$ represents the graph with adjacency matrix elements independently sampled from the Bernoulli distribution with an equal probability of having an edge or not, which is equivalent to sampling from
Erd{\H{o}}s-R{\'e}nyi random graphs \cite{erdHos1960evolution} with the edge probability of $0.5$.
From the perspective of $\mathbf{\bar{A}}_t$, it is actually a diffusion process on the discrete distribution. 
We emphasize the significance of leveraging the association with discrete graphs, which allows us to gain additional observations of intermediate graph states, such as changes in graph structure and position, while still benefiting from the convenient continuous SDE descriptions.

With the defined forward diffusion process including both continuous and discrete states, we build up the training procedure shown in Fig. \ref{fig:train} to supervise the learning of the graph score network.
The training objective is defined as 
\begin{equation}
\begin{split}
    \min_{\mathbf{\theta}}\mathbb{E}_{t}\{
    \lambda(t) \mathbb{E}_{\mathbf{A}_0} \mathbb{E}_{\mathbf{A}_t | \mathbf{A}_0}
    [\,||\mathbf{s_\theta}(\mathbf{A}_t, \mathbf{\bar{A}}_t , t) \\ - \nabla_{\mathbf{A}}\mathrm{log}\ p_{0t}(\mathbf{A}_t|\mathbf{A}_0)
    ||_2^2\,]
    \}.
\end{split} 
\label{eq:graphloss}
\end{equation}
We expect to utilize hybrid states for the graph score network to denoise the added Gaussian noise.
At the beginning of diffusion processes, the node position, including the distance to other nodes, is quickly lost in real-valued adjacency matrices, since every element in the adjacency matrix is perturbed at the same time.
Thus, if we estimate scores directly on dense adjacency matrices like EDP-GNN \cite{niu2020permutation}, such graph topology information will be absent, which is an important clue for graph structure representation learning.
This motivates us to better design the graph score network.

\subsection{Position-enhanced Graph Score Networks} \label{subsec:pgsn}
A graph score network $\mathbf{s}_{\theta}(\mathbf{A}_t, \mathbf{\bar{A}}_t , t)$ learns to estimate the score of graphs with varying perturbation degrees, also known as a denoising model.
Message passing based graph neural networks (GNNs) 
have become the de facto standard for graph structure representation learning, whose representation power is bounded by
1-WL test \cite{weisfeiler1968reduction, xu2018powerful, morris2021weisfeiler}. 
To enhance the representation power of GNNs, recent works \cite{you2019position, dwivedi2020generalization, li2020distance, kreuzer2021rethinking, dwivedi2021lspe} attempt to add the position encoding of nodes or edges to message passing architectures, including the position information from graph spectrum, random walks, and so on.
Here, we propose a specialized \textbf{P}osition-enhanced \textbf{G}raph \textbf{S}core \textbf{N}etwork (PGSN) that utilizes graph structure and position features from intermediate discrete graphs, with real-valued adjacency matrices serving as initial edge features and contributing to score estimation through an edge feature updating mechanism. 

We first extract appropriate node features and edge features from $\mathbf{A}_t$ and $ \mathbf{\bar{A}}_t$.
As the time information is added to all extracted features with the sinusoidal position embedding \cite{vaswani2017attention}, we omit $t$ in the description of PGSN.
We take the degree onehot feature $\mathbf{h}^0$ as the initialization of node features, and obtain graph position information via the random walk $\mathrm{RW} = \mathbf{\bar{A}}\mathbf{D}^{-1}$ following \cite{li2020distance, dwivedi2021lspe}.
The node position feature consists of the landing probabilities of node $i$ from the $r$-step random walks are defined as:
\begin{equation}
    \mathbf{p}_i = [\,\mathrm{RW}_{ii}, \mathrm{RW}_{ii}^2, \cdots, \mathrm{RW}_{ii}^r \,] \ .
\end{equation}
Using the same matrix of random walks, we also obtain the shortest-path-distance feature between node $i$ and node $j$ by
\begin{equation}
    \mathbf{e}_{ij}^{spd} = \phi([\,\mathrm{RW}_{ij}, \mathrm{RW}_{ij}^2, \cdots, \mathrm{RW}_{ij}^r \,]) \ ,
\end{equation}
where $\phi(\cdot)$ takes the first non-zero position of the input vector and turns it to the onehot feature.
Then the initial edge features are concatenated by $\mathbf{e}^0 = [\mathbf{A}W_0, \mathbf{e}^{spd}] $.

In order to capture the current graph structure thoroughly, we apply an $L$-layer message passing architecture incorporating node and edge features.
The edge set is constructed by elements in $\mathbf{A}$ that are greater than the threshold $\gamma$ (a hyperparameter controlling the computation burden and usually set to 0.2 in our experiments).
We compute the two types of message for the node $i$ at the $l$-th message passing layer as follows:
\begin{gather}
	\alpha_{i,j}^{k,l} = \mathrm{softmax}\left(\frac{\mathbf{q}_{i}^{k,l} \  (\mathbf{k}_j^{k,l}\circ \mathbf{c}_{i,j}^{k,l})^{\top}}{\sqrt{d}} \right) \ , \\
	\mathbf{m}_{i,j, (h)}^{k, l} = \alpha_{i,j}^{k,l} \ \mathbf{v}_j^{k,l} \circ \mathbf{\bar{c}}_{i,j}^{k,l} \ , \\
    \mathbf{m}_{i,j, (p)}^{k, l} = \mathbf{m}_{i,j, (h)}^{k, l} \circ \mathbf{p}_j^{l} W_p^{k,l} \ , 
\end{gather}
where $\circ$ denotes element-wise multiplication.
Here, $\mathbf{q}^l$ , $\mathbf{k}^l$ and $\mathbf{v}^l$ are node features projected by different learnable matrices from the concatenation $[\mathbf{h}^l, \mathbf{p}^l]$, while $\mathbf{c}^{l}$ and $\mathbf{\bar{c}}^l$ are edge features projected from $\mathbf{e}^l$.
The edge features not only bias the attention computation but also as a part of the aggregated features.
We aggregate and update the node embedding and the node position encoding by
\begin{gather}
\mathbf{M}_{i, (h)}^{l} = \mathop {||}\limits_{k = 1}^H \sum_{j \in N(i)} \mathbf{m}_{i,j, (h)}^{k, l} \ , \\
\hat{\mathbf{h}}_i^{l+1} = \mathrm{Norm}(\mathbf{M}_{i, (h)}^{l} + \mathbf{h}_i^l W_1 ) \ , \\
\mathbf{h}_i^{l+1} = \mathrm{Norm}\left(\hat{\mathbf{h}}_i^{l+1} + \mathrm{FFN}(\hat{\mathbf{h}}_i^{l+1})\right) \ , \\
\mathbf{M}_{i, (p)}^{l} = \mathop {||}\limits_{k = 1}^H \sum_{j \in N(i)} \mathbf{m}_{i,j, (p)}^{k, l} \ , \\
\mathbf{p}_i^{l+1} = \mathbf{p}_i^{l} + \mathrm{act}(\mathbf{M}_{i, (p)}^{l} + \mathbf{p}_i^{l}) \ ,
\end{gather}
where $||$ is the concatenation of multi-head messages, $\mathrm{Norm}$ is a normalization layer, $\mathrm{FFN}$ is a two-layer feed forward network and $\mathrm{act}$ is an activation layer.
Another important step is to update the edge feature after message passing as 
\begin{equation}
\mathbf{e}_{i,j}^{l+1} = \mathbf{e}_{i,j}^l + \mathrm{act}\left((\mathbf{h}_i^{l+1}+\mathbf{h}_j^{l+1})W_2\right) .
\end{equation}
Getting the final edge embedding $\mathbf{e}^{L}$, we concatenate it with the original $\mathbf{e}^0$ and adopt a multilayer perceptron (MLP) for the score estimation of each edge.

The standard message passing in graphs is theoretically guaranteed to be permutation equivariant \cite{ michael21geometric}.
Since the extracted node features and edge features are permutation equivariant, 
and since the operations in our PGSN consist purely of message passing and node-wise/edge-wise projections, 
the output edge scores are still permutation equivariant.

\subsection{Graph Generation via Generative Diffusion Processes} \label{subsec:graphgdp}

After training with the time-dependent graph score network $\mathbf{s_{\theta}}$, we can construct the generative diffusion processes by the reverse-time SDE from Eq. (\ref{eq:reverse_sde}) as
\begin{equation}
\mathrm{d}\mathbf{A} = [\,- \frac{1}{2} \beta(t) \mathbf{A}  -  \beta(t)\nabla_{\mathbf{A}}\mathrm{log}\ p_{t}(\mathbf{A})\,]\mathrm{d}t +
\sqrt{\beta(t)}\mathrm{d}\bar{\mathbf{w}} \ , 
\label{eq:graph_reverse}
\end{equation}
where $\nabla_{\mathbf{A}}\mathrm{log}\ p_{t}(\mathbf{A})$ is the score function parametrized by $\mathbf{s_{\theta}}$.
By first sampling from the prior Normal distribution (\ie sampling from random graphs), a new graph instance is generated by performing the generative diffusion process described by Eq. (\ref{eq:graph_reverse}). 
Therefore, graph generation is transformed into a problem of numerically solving SDEs.

We utilize three numerical methods for solving the special reverse-time SDE, suitable for graph generation in different situations.
First, many numerical solvers directly provide the approximate trajectory simulation of SDEs.
For example, the Euler-Maruyama method is a simple discretization to the SDE, which is defined as 
\begin{small}
\begin{equation}
\mathbf{A}_{t-\Delta t} = \mathbf{A}_{t} + [\frac{1}{2} \beta(t) \mathbf{A}_{t}  +  \beta(t)\mathbf{s_{\theta}} (\mathbf{A}_t, \bar{\mathbf{A}}_t, t)]\Delta t
+ \sqrt{\beta(t)} \sqrt{\Delta t}\mathbf{z}_t, 
\label{eq:EM}
\end{equation}
\end{small}
where $\mathbf{z}_t \sim \mathcal{N}(\mathbf{0}, \mathbf{I})$.
Given the number of discretization steps, we can determine $\Delta t$ and generate graphs iteratively using the gradient information from the graph score network in Eq. (\ref{eq:EM}). 
For simple and small graphs, such general SDE solvers provide adequate generation quality.

Second, for those graphs with complex structural characteristics (\eg, large graph diameters), we can further employ Langevin MCMC \cite{parisi1981correlation} like score-based models \cite{song2019generative, song2021score} to improve the sample quality at each discretization step.
Specifically, after using Eq.\ref{eq:EM} to estimate the graph sample for the next step, we correct the estimated graph sample by
\begin{equation}
    \mathbf{A}_{t} \leftarrow \mathbf{A}_{t} + \epsilon_{t}\mathbf{s_{\theta}} (\mathbf{A}_t, \bar{\mathbf{A}}_t, t) + \sqrt{2\epsilon_{t}} \mathbf{z} \ , 
\end{equation}
where the step size $\epsilon_{t}$ is determined by the norm of noise, the norm of scores and a hyperparameter $r$.
The extra correction steps reduces the error from numerical solvers, obtaining more accurate margin distribution of graph samples.

Third, we can employ the corresponding ODE for efficient graph generation.
\cite{SongDDIM21} and \cite{song2021score} have explored the connection between generative diffusion processes and ODEs. Without the stochastic term in Eq. (\ref{eq:graph_reverse}), we can derive the probability flow ODE \cite{song2021score}, which corresponds to a deterministic process sharing the same marginal probability densities with the SDE.
The probability flow ODE is defined as
\begin{equation}
\mathrm{d}\mathbf{A} = [\,- \frac{1}{2} \beta(t) \mathbf{A}  -  \frac{1}{2}\beta(t)\mathbf{s_{\theta}} (\mathbf{A}, \bar{\mathbf{A}}, t)\,]\mathrm{d}t \ .
\label{eq:ode}
\end{equation}
It allows us to use current well-established ODE solvers to generate high-quality graphs in very few steps, which we describe in detail in Section \ref{subsec:tradeoff}.
We quantize all the generated continuous adjacency matrices for final graphs.

\section{Experiments} \label{section4}
In this section, we empirically demonstrate the power of the proposed GraphGDP in the task of graph generation. 

\subsection{Datasets}
We compare our graph generative model on four common graph datasets that vary in graph sizes and characteristics. 
(1) \textit{Community-small}: 100 community graphs with $12 \leq|V| \leq 20$. The graphs are constructed by two communities with equal nodes, each of which is generated by the Erd{\H{o}}s-R{\'e}nyi model (E-R) \cite{erdHos1960evolution} with $p=0.7$. The inter-community edges are added with the uniform probability $0.05$. 
(2) \textit{Ego-small}: 200 one-hop ego graphs with  $4 \leq |V| \leq 18$, extracted from Citeseer network \cite{sen2008collective}. The nodes represent documents and edges represent citation relationships.
(3) \textit{Ego}: 757 three-hop ego graphs with $50 \leq |V| \leq 399$, also extracted from Citeseer network \cite{sen2008collective}.
(4) \textit{Enzymes}: 563 protein graphs with $10 \leq |V| \leq 125$ selected from BRENDA database \cite{schomburg2004brenda}.

We further split the datasets into training and test sets with a ratio of $8:2$. The validation set comes from the first $20\%$ of the training graphs.
When evaluating the model performance on Community-small and Ego-small, we generate $1024$ graph samples following \cite{nips/LiuKBKS19, niu2020permutation} to receive more stable evaluation results on small graphs.
For Ego and Enzymes, we generate the same number of graphs as the test set.

\subsection{Evaluation Metrics}
Evaluating and comparing graph generative models is a challenging task, as it is difficult to obtain perceptual differences for graph visualization.
We apply two type of metrics to comprehensively evaluate the quality of graph generation.

\subsubsection{Classical Structure Metrics}
The widely-used evaluation metrics are based on Maximum Mean Discrepancy (MMD) measures to assess the distance between the distributions of the generated graph set $\mathbb{S}_{g}$ and the test set $\mathbb{S}_{t}$ \cite{you2018graphrnn, nips/LiaoLSWHDUZ19, nips/LiuKBKS19, niu2020permutation, dai2020scalable, icml/ChenHHRL21}.
Specifically, several graph property descriptor functions (\eg degree distribution, clustering coefficient, 4-node orbit count histograms, and Laplacian spectrum) are applied to map each graph to high-dimensional representations.
The estimate of MMD \cite{gretton2012kernel} on these representations can be derived as 
\begin{equation}
\begin{aligned}
\mathbf{MMD}(\mathbb{S}_{g},\mathbb{S}_{t}) &:= \frac{1}{m^2} \sum_{i,j=1}^{m}k(\mathbf{x}_i^{t}, \mathbf{x}_j^{t}) + \frac{1}{n^2}\sum_{i,j=1}^{n}k(\mathbf{x}_i^{g}, \mathbf{x}_j^{g}) \\
& - \frac{2}{nm}\sum_{i=1}^{n}\sum_{j=1}^{m}k(\mathbf{x}_i^{g}, \mathbf{x}_{j}^{t}) \ ,     
\end{aligned}
\end{equation}
where $k(\cdot , \cdot)$ is an optional kernel function, including a kernel using the first Wasserstein distance (EMD) or total variation distance (TV), and the radial basis function kernel (RBF). 

Recently, O'Bray \etal \cite{o2021evaluation} point out that the current practice of MMD metrics fail to faithfully reflect the distance of graph distributions.
For example, MMD necessitates the use of positive definite kernel functions, and the previously-used hyperparameters fail to align with maximum discrimination in MMD.
Following the suggestions from \cite{o2021evaluation}, we employ a more reasonable structure evaluation process to reflect the performance of graph generative models.
\textbf{First}, we choose a valid and efficient kernel function, \ie an RBF kernel with a smoothing parameter $\sigma \in \mathbb{R}$ as $k(x_{i}, x_{j}) = exp(\frac{{-\|x_i-x_j\|}^2}{2\sigma^2})$.
\textbf{Second}, we employ three graph-level structure descriptor functions, which are described in \cite{you2018graphrnn, nips/LiaoLSWHDUZ19}, including (i) the degree distribution, (ii) the clustering coefficient distribution, and (iii) the Laplacian spectrum histograms.
\textbf{Third}, we report the highest MMD values under a set of $\sigma$, expected to show the maximum distance between the two distributions. 
The $50$ candidate values of $log \sigma$ are taken evenly at the interval in $[10^{-5}, 10^{5}]$. 
As for the number of bins used for the histogram conversion, we inherit the setting of \cite{you2018graphrnn, nips/LiaoLSWHDUZ19}, which takes $100$ bins for clustering coefficient and $200$ bins for Laplacian spectrum.
\textbf{Four}, MMD between the test and training graphs is included to provide a meaningful performance bound. 

\subsubsection{Neural-network-based Metrics}
Thompson \etal \cite{thompson2022Metric} introduce several random GIN-based metrics for graph generative model evaluation, as the pre-existing structure metrics fail to capture the diversity of graph samples.
The graph representations are extracted by random-initialized GIN \cite{xu2018powerful}, where MMD RBF (\ie MMD computed with the RBF kernel), F1 PR (\ie the harmonic mean of improved precision and recall) and F1 DC (\ie the harmonic mean of density and coverage) are built.
MMD RBF is a more stable and comprehensive metric to measure the diversity and realism of generated graphs, while F1 PR and F1 DC are sensitive to detecting mode collapse and mode dropping. We follow the configuration of GIN in \cite{thompson2022Metric} and report the mean result of $10$ random GINs.

\begin{table*}[t]
\caption{Comparison of the graph generation performance among graph generative models with classical structure metrics. The Train/Test shows the MMD results in a set of $\sigma$ values between training and test graphs.
For MMD metrics, the closer the value is to the Train/Test results, the better the performance. The top two cells in each column are coloured according to their rank. 
Deg.: degree distribution, Clus.: clustering coefficient distribution, Spec.: spectrum of graph Laplacian, Avg.: the average values of three MMD metrics.}
\label{tab:main}
\centering
\resizebox{\textwidth}{!}{%
\renewcommand{\arraystretch}{1.25}
\begin{tabular}{llcccclcccclcccclcccc}
\hline
 &
   &
  \multicolumn{4}{c}{Community-small} &
   &
  \multicolumn{4}{c}{Ego-small} &
   &
  \multicolumn{4}{c}{Enzymes} &
   &
  \multicolumn{4}{c}{Ego} \\ \cline{3-6} \cline{8-11} \cline{13-16} \cline{18-21} 
 &
   &
  \multicolumn{4}{c}{$|V|_{max}=20, |E|_{max}=62$} &
   &
  \multicolumn{4}{c}{$|V|_{max}=17, |E|_{max}=66$} &
   &
  \multicolumn{4}{c}{$|V|_{max}=125, |E|_{max}=149$} &
   &
  \multicolumn{4}{c}{$|V|_{max}=399, |E|_{max}=1071$} \\
 &
   &
  \multicolumn{4}{c}{$|V|_{avg}\approx15, |E|_{avg}\approx36$} &
   &
  \multicolumn{4}{c}{$|V|_{avg}\approx6, |E|_{avg}\approx9$} &
   &
  \multicolumn{4}{c}{$|V|_{avg}\approx33, |E|_{avg}\approx63$} &
   &
  \multicolumn{4}{c}{$|V|_{avg}\approx145, |E|_{avg}\approx335$} \\ \cline{3-6} \cline{8-11} \cline{13-16} \cline{18-21} 
 &
   &
  Deg. &
  Clus. &
  Spec. &
  Avg. &
   &
  Deg. &
  Clus. &
  Spec. &
  Avg. &
   &
  Deg. &
  Clus. &
  Spec. &
  Avg. &
   &
  Deg. &
  Clus. &
  Spec. &
  Avg. \\ \hline
Train/Test &
   &
  \textbf{0.035} &
  \textbf{0.067} &
  \textbf{0.045} &
  \textbf{0.049} &
   &
  \textbf{0.025} &
  \textbf{0.029} &
  \textbf{0.027} &
  \textbf{0.027} &
   &
  \textbf{0.011} &
  \textbf{0.011} &
  \textbf{0.011} &
  \textbf{0.011} &
   &
  \textbf{0.009} &
  \textbf{0.009} &
  \textbf{0.009} &
  \textbf{0.009} \\ \hline
\textbf{Order-dependent} &
   &
  \multicolumn{1}{l}{} &
  \multicolumn{1}{l}{} &
  \multicolumn{1}{l}{} &
  \multicolumn{1}{l}{} &
   &
  \multicolumn{1}{l}{} &
  \multicolumn{1}{l}{} &
  \multicolumn{1}{l}{} &
  \multicolumn{1}{l}{} &
   &
  \multicolumn{1}{l}{} &
  \multicolumn{1}{l}{} &
  \multicolumn{1}{l}{} &
  \multicolumn{1}{l}{} &
   &
  \multicolumn{1}{l}{} &
  \multicolumn{1}{l}{} &
  \multicolumn{1}{l}{} &
  \multicolumn{1}{l}{} \\ \hline
GraphRNN &
   &
  0.106 &
  0.115 &
  0.091 &
  0.104 &
   &
  0.155 &
  0.229 &
  0.167 &
  0.184 &
   &
  0.397 &
  0.302 &
  0.260 &
  0.320 &
   &
  0.140 &
  0.755 &
  0.316 &
  0.404 \\
GRAN &
   &
  0.125 &
  0.164 &
  0.111 &
  0.133 &
   &
  0.096 &
  0.072 &
  0.095 &
  0.088 &
   &
  0.215 &
  0.147 &
  0.034 &
  0.132 &
   &
  0.594 &
  0.425 &
  1.025 &
  0.682 \\
BIGG &
   &
  \cellcolor[HTML]{D6FDD0}0.041 &
  \cellcolor[HTML]{92FA7A}0.073 &
  \cellcolor[HTML]{92FA7A}0.050 &
  \cellcolor[HTML]{92FA7A}0.055 &
   &
  \cellcolor[HTML]{92FA7A}0.024 &
  \cellcolor[HTML]{92FA7A}0.029 &
  \cellcolor[HTML]{92FA7A}0.028 &
  \cellcolor[HTML]{92FA7A}0.027 &
   &
  \cellcolor[HTML]{92FA7A}0.020 &
  \cellcolor[HTML]{92FA7A}0.019 &
  \cellcolor[HTML]{92FA7A}0.019 &
  \cellcolor[HTML]{92FA7A}0.019 &
   &
  \cellcolor[HTML]{92FA7A}0.034 &
  \cellcolor[HTML]{D6FDD0}0.108 &
  \cellcolor[HTML]{D6FDD0}0.077 &
  \cellcolor[HTML]{D6FDD0}0.073 \\ \hline
\textbf{Order-independent} &
   &
  \multicolumn{1}{l}{} &
  \multicolumn{1}{l}{} &
  \multicolumn{1}{l}{} &
  \multicolumn{1}{l}{} &
   &
  \multicolumn{1}{l}{} &
  \multicolumn{1}{l}{} &
  \multicolumn{1}{l}{} &
  \multicolumn{1}{l}{} &
   &
  \multicolumn{1}{l}{} &
  \multicolumn{1}{l}{} &
  \multicolumn{1}{l}{} &
  \multicolumn{1}{l}{} &
   &
  \multicolumn{1}{l}{} &
  \multicolumn{1}{l}{} &
  \multicolumn{1}{l}{} &
  \multicolumn{1}{l}{} \\ \hline
ER &
   &
  0.300 &
  0.239 &
  0.100 &
  0.213 &
   &
  0.200 &
  0.094 &
  0.361 &
  0.218 &
   &
  0.844 &
  0.381 &
  0.104 &
  0.443 &
   &
  0.738 &
  0.397 &
  0.868 &
  0.668 \\
VGAE &
   &
  0.391 &
  0.257 &
  0.095 &
  0.248 &
   &
  0.146 &
  0.046 &
  0.249 &
  0.147 &
   &
  0.811 &
  0.514 &
  0.153 &
  0.493 &
   &
  0.873 &
  1.210 &
  0.935 &
  1.006 \\
GraphRNN-U &
   &
  0.410 &
  0.297 &
  0.103 &
  0.270 &
   &
  0.471 &
  0.416 &
  0.398 &
  0.429 &
   &
  0.932 &
  1.000 &
  0.367 &
  0.766 &
   &
  1.413 &
  1.097 &
  1.110 &
  1.207 \\
GRAN-U &
   &
  0.106 &
  0.127 &
  0.083 &
  0.106 &
   &
  0.155 &
  0.229 &
  0.167 &
  0.184 &
   &
  0.343 &
  0.122 &
  0.041 &
  0.169 &
   &
  0.099 &
  0.170 &
  0.179 &
  0.149 \\
EDP-GNN &
   &
  0.100 &
  0.140 &
  0.085 &
  0.108 &
   &
  \cellcolor[HTML]{92FA7A}{\color[HTML]{000000} 0.026} &
  0.032 &
  0.037 &
  0.032 &
   &
  0.120 &
  0.644 &
  0.070 &
  0.278 &
   &
  0.553 &
  0.605 &
  0.374 &
  0.511 \\
GraphGDP &
   &
  \cellcolor[HTML]{92FA7A}0.039 &
  \cellcolor[HTML]{D6FDD0}0.074 &
  \cellcolor[HTML]{D6FDD0}0.052 &
  \cellcolor[HTML]{92FA7A}0.055 &
   &
  0.023 &
  \cellcolor[HTML]{92FA7A}0.029 &
  \cellcolor[HTML]{D6FDD0}0.030 &
  \cellcolor[HTML]{92FA7A}0.027 &
   &
  \cellcolor[HTML]{D6FDD0}0.023 &
  \cellcolor[HTML]{D6FDD0}0.025 &
  \cellcolor[HTML]{92FA7A}0.019 &
  \cellcolor[HTML]{D6FDD0}0.022 &
   &
  \cellcolor[HTML]{D6FDD0}0.037 &
  \cellcolor[HTML]{92FA7A}0.099 &
  \cellcolor[HTML]{92FA7A}0.021 &
  \cellcolor[HTML]{92FA7A}0.052 \\ \hline
\end{tabular}
}
\end{table*}

\begin{table*}[!htbp]
\caption{Evaluation of different graph generative models using three neural-network-based metrics.
The $50/50$ split represents the results computed with a random $50/50$ split of the dataset and  shows the ideal scores for metrics.
The top two cells in each column are coloured according to their rank.}
\label{tab:nn-based}
\centering
\resizebox{\textwidth}{!}{%
\renewcommand{\arraystretch}{1.45}
\begin{tabular}{lccccccccccc}
\hline
 &
  \multicolumn{3}{c}{Community-small} &
   &
  \multicolumn{3}{c}{Enzymes} &
   &
  \multicolumn{3}{c}{Ego} \\
 &
  MMD RBF ($\downarrow$) &
  F1 PR ($\uparrow$) &
  F1 DC ($\uparrow$) &
   &
  MMD RBF ($\downarrow$) &
  F1 PR ($\uparrow$) &
  F1 DC ($\uparrow$) &
   &
  MMD RBF ($\downarrow$) &
  F1 PR ($\uparrow$) &
  F1 DC ($\uparrow$) \\ \hline
50/50 split &
  \textbf{0.037 $\pm$ 0.002} &
  \textbf{0.994 $\pm$ 0.012} &
  \textbf{1.065 $\pm$ 0.008} &
   &
  \textbf{0.007 $\pm$ 0.000} &
  \textbf{0.988 $\pm$ 0.004} &
  \textbf{0.979 $\pm$ 0.006} &
   &
  \textbf{0.005 $\pm$ 0.000} &
  \textbf{0.985 $\pm$ 0.004} &
  \textbf{1.025 $\pm$ 0.012} \\ \hline
\textbf{Order-dependent} &
   &
   &
   &
   &
   &
   &
   &
   &
   &
   &
   \\ \hline
GraphRNN &
  0.353 $\pm$ 0.088 &
  0.252 $\pm$ 0.183 &
  0.407 $\pm$ 0.171 &
   &
  1.495 $\pm$ 0.037 &
  0.000 $\pm$ 0.000 &
  0.000 $\pm$ 0.000 &
   &
  1.283 $\pm$ 0.053 &
  0.019 $\pm$ 0.016 &
  0.007 $\pm$ 0.007 \\
GRAN &
  0.196 $\pm$ 0.014 &
  0.824 $\pm$ 0.141 &
  0.793 $\pm$ 0.099 &
   &
  0.069 $\pm$ 0.008 &
  0.915 $\pm$ 0.035 &
  0.738 $\pm$ 0.027 &
   &
  0.244 $\pm$ 0.064 &
  0.238 $\pm$ 0.141 &
  0.207 $\pm$ 0.088 \\
BIGG &
  \cellcolor[HTML]{92FA7A}0.052 $\pm$ 0.003 &
  0.135 $\pm$ 0.087 &
  \cellcolor[HTML]{92FA7A}1.048 $\pm$ 0.035 &
   &
  \cellcolor[HTML]{92FA7A}0.019 $\pm$ 0.000 &
  \cellcolor[HTML]{D6FDD0}0.964 $\pm$ 0.008 &
  \cellcolor[HTML]{92FA7A}0.966 $\pm$ 0.012 &
   &
  \cellcolor[HTML]{92FA7A}0.022 $\pm$ 0.002 &
  \cellcolor[HTML]{92FA7A}0.956 $\pm$ 0.014 &
  \cellcolor[HTML]{92FA7A}0.896 $\pm$ 0.026 \\ \hline
\textbf{Order-independent} &
   &
   &
   &
   &
   &
   &
   &
   &
   &
   &
   \\ \hline
ER &
  0.278 $\pm$ 0.046 &
  0.363 $\pm$ 0.201 &
  0.335 $\pm$ 0.096 &
   &
  0.808 $\pm$ 0.065 &
  0.046 $\pm$ 0.030 &
  0.019 $\pm$ 0.005 &
   &
  0.118 $\pm$ 0.035 &
  0.516 $\pm$ 0.116 &
  0.377 $\pm$ 0.120 \\
VGAE &
  0.360 $\pm$ 0.065 &
  0.292 $\pm$ 0.165 &
  0.292 $\pm$ 0.113 &
   &
  0.716 $\pm$ 0.033 &
  0.012 $\pm$ 0.016 &
  0.002 $\pm$ 0.003 &
   &
  0.520 $\pm$ 0.003 &
  0.000 $\pm$ 0.000 &
  0.000 $\pm$ 0.000 \\
GraphRNN-U &
  0.970 $\pm$ 0.113 &
  0.066 $\pm$ 0.043 &
  0.079 $\pm$ 0.003 &
   &
  1.263 $\pm$ 0.177 &
  0.000 $\pm$ 0.000 &
  0.000 $\pm$ 0.000 &
   &
  1.317 $\pm$ 0.022 &
  0.000 $\pm$ 0.000 &
  0.000 $\pm$ 0.001 \\
GRAN-U &
  0.164 $\pm$ 0.016 &
  \cellcolor[HTML]{D6FDD0}0.859 $\pm$ 0.082 &
  0.888 $\pm$ 0.053 &
   &
  0.242 $\pm$ 0.033 &
  0.671 $\pm$ 0.056 &
  0.364 $\pm$ 0.024 &
   &
  0.128 $\pm$ 0.041 &
  0.720 $\pm$ 0.041 &
  0.564 $\pm$ 0.026 \\
EDP-GNN &
  0.125 $\pm$ 0.004 &
  \cellcolor[HTML]{92FA7A}0.913 $\pm$ 0.108 &
  0.977 $\pm$ 0.044 &
   &
  0.119 $\pm$ 0.010 &
  0.954 $\pm$ 0.012 &
  0.846 $\pm$ 0.020 &
   &
  0.295 $\pm$ 0.061 &
  0.395 $\pm$ 0.028 &
  0.192 $\pm$ 0.036 \\
GraphGDP &
  \cellcolor[HTML]{D6FDD0}0.066 $\pm$ 0.012 &
  0.656 $\pm$ 0.138 &
  \cellcolor[HTML]{D6FDD0}1.042 $\pm$ 0.014 &
   &
  \cellcolor[HTML]{D6FDD0}0.026 $\pm$ 0.001 &
  \cellcolor[HTML]{92FA7A}0.974 $\pm$ 0.005 &
  \cellcolor[HTML]{D6FDD0}0.932 $\pm$ 0.015 &
   &
  \cellcolor[HTML]{D6FDD0}0.034 $\pm$ 0.004 &
  \cellcolor[HTML]{D6FDD0}0.877 $\pm$ 0.014 &
  \cellcolor[HTML]{D6FDD0}0.721 $\pm$ 0.023 \\ \hline
\end{tabular}%
}
\end{table*}

\begin{figure*}[t]
    \centering
    \adjustimage{width=0.93\textwidth, center}{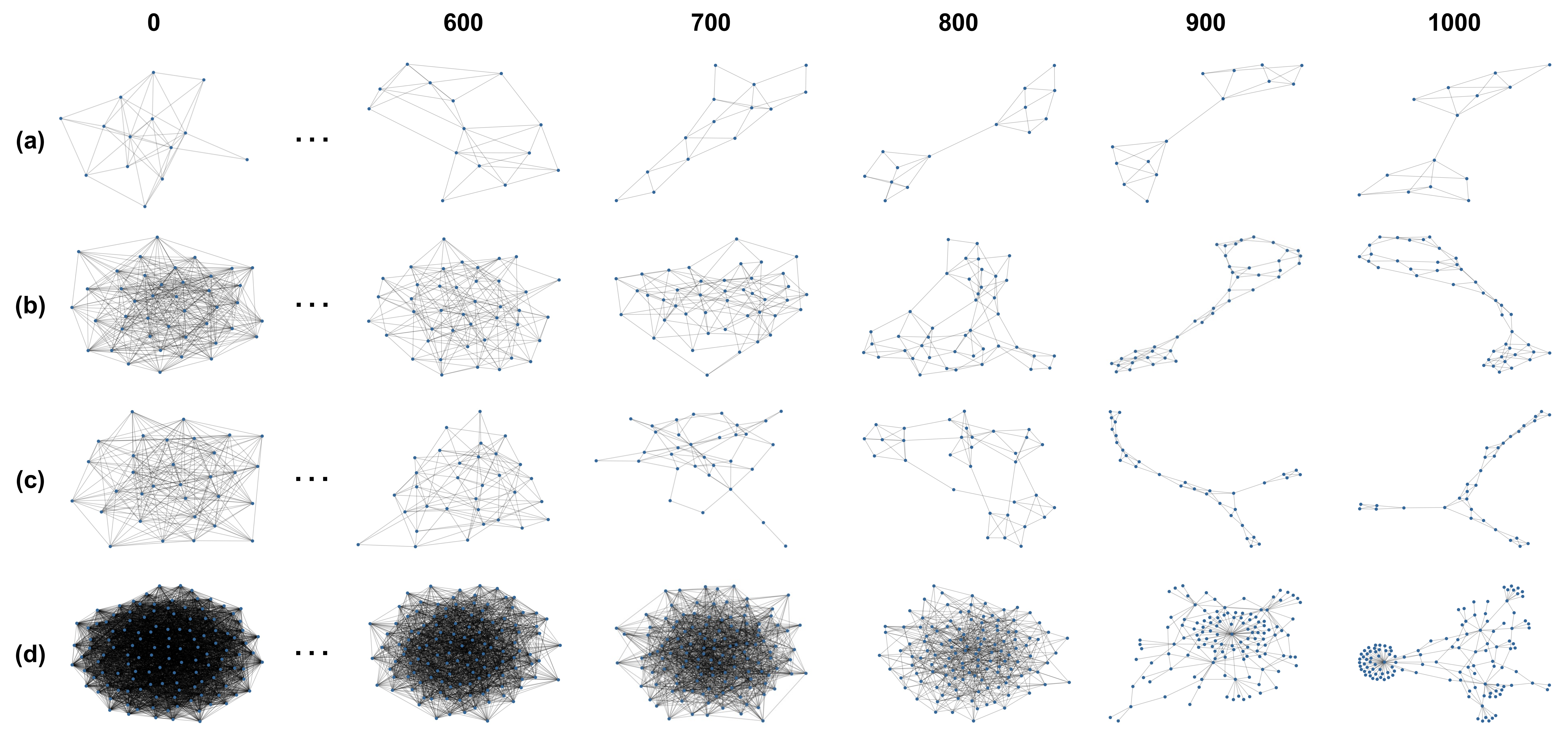}
    \caption{Graph visualization of different steps in the generative diffusion processes on Community-small (a), Enzymes (b-c), Ego (d) datasets.}
    \label{fig:graph_vis}
\end{figure*}

\subsection{Baselines}
We compare the performance of our models against other graph generative models including VGAE \cite{kipf2016variational}, GraphRNN \cite{you2018graphrnn}, GRAN \cite{nips/LiaoLSWHDUZ19}, EDP-GNN \cite{niu2020permutation} and BIGG \cite{dai2020scalable}.
For autoregressive graph generative models, we train models with the breadth-first-search (BFS) or depth-first-search (DFS) canonical node ordering schemes.
In addition, we utilize uniformly distributed random node orderings to train several autoregressive models denoted by the extra $\mathbf{-U}$, which can be considered as an order-agnostic autoregressive model \cite{uria2014deep} that maximize the average likelihood over all node orderings of the graph \cite{li2018learning,icml/ChenHHRL21}.
An Erd{\H{o}}s-R{\'e}nyi (\textbf{ER}) baseline \cite{erdHos1960evolution} is also added, where the edge probability is estimated by the maximum likelihood over training graphs.
The brief explanations and implementation details of deep graph generative models are listed as follows.
\textbf{VGAE} \cite{kipf2016variational} is a variational autoencoder that utilizes a graph convolution network encoder and a simple MLP decoder with inner product. 
\textbf{GraphRNN} \cite{you2018graphrnn} is an autoregressive model using a graph-level Recurrent Neural Network (RNN) to maintain graph states and another edge-level RNN to generate edges of the newly generated node. 
We re-train the officially implemented models with our dataset split using random BFS node orderings.
\textbf{GRAN} \cite{nips/LiaoLSWHDUZ19} maintains the autoregressive process and utilizes graph neural networks with attention to model the graph generated context.
GRAN generates a row of the adjacency matrix in a decision step to improve efficiency. 
The fixed DFS node orderings are utilized for Enzymes dataset, and the BFS node orderings for others. 
\textbf{BIGG} \cite{dai2020scalable} is the state-of-the-art autoregressive tree-based model which utilizes the sparsity of realistic graphs. 
Over the sequence of nodes, it adopts a binary tree data structure to generate each edge and associates the set of edges with each node via a tree-structured autoregressive model.
\textbf{EDP-GNN} \cite{niu2020permutation} is a permutation invariance approach for graph generation via graph score matching and annealed Langevin dynamic sampling.
We exploit its original model hyperparameters.

\begin{figure}[t]
    \centering
    \begin{subfigure}{\columnwidth}
        \centering
        \includegraphics[width=\columnwidth]{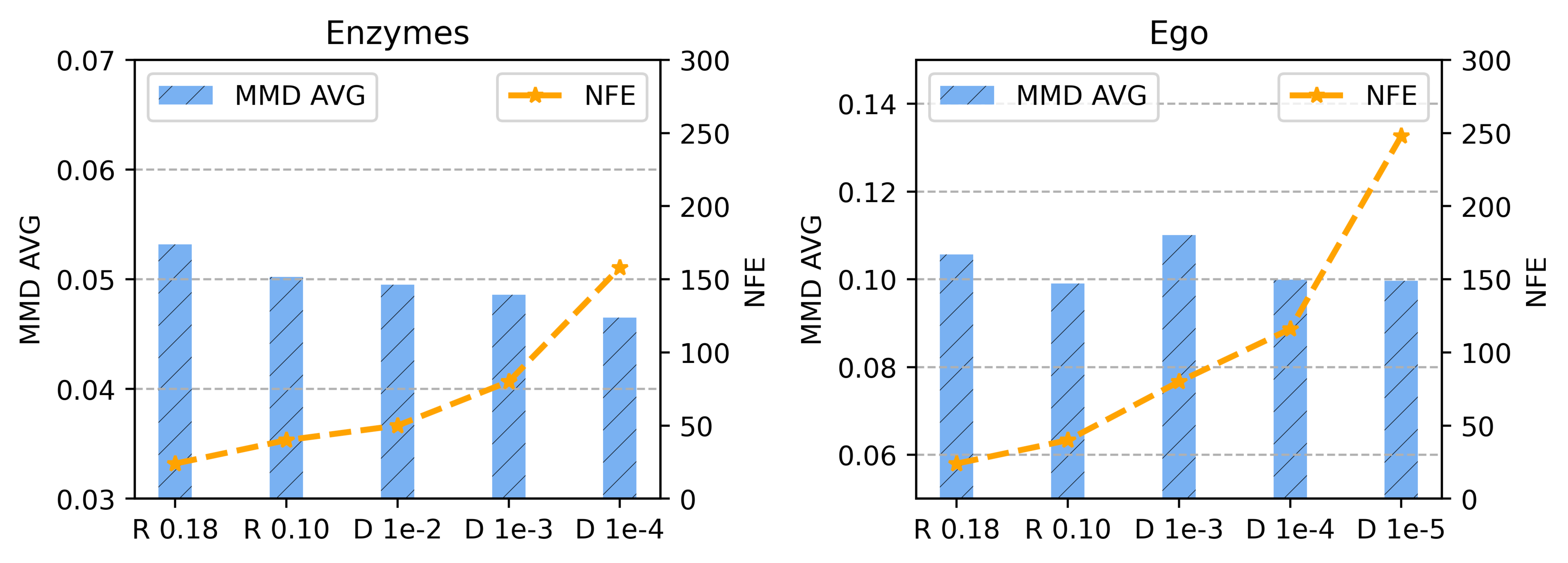}
        \caption{Graph generation performance and NFE used by ODE solvers. 
        MMD AVG: average values of three structure MMD metrics.
        R 0.18: "rk4" fixed-step solver with 0.18 step size.
        D 1e-2: "dopri5" adaptive-step solver with 1e-2 error tolerance. }
        \label{subfig:ode-res}
    \end{subfigure}
    \begin{subfigure}{\columnwidth}
        \centering
        \adjustimage{width=\columnwidth, center}{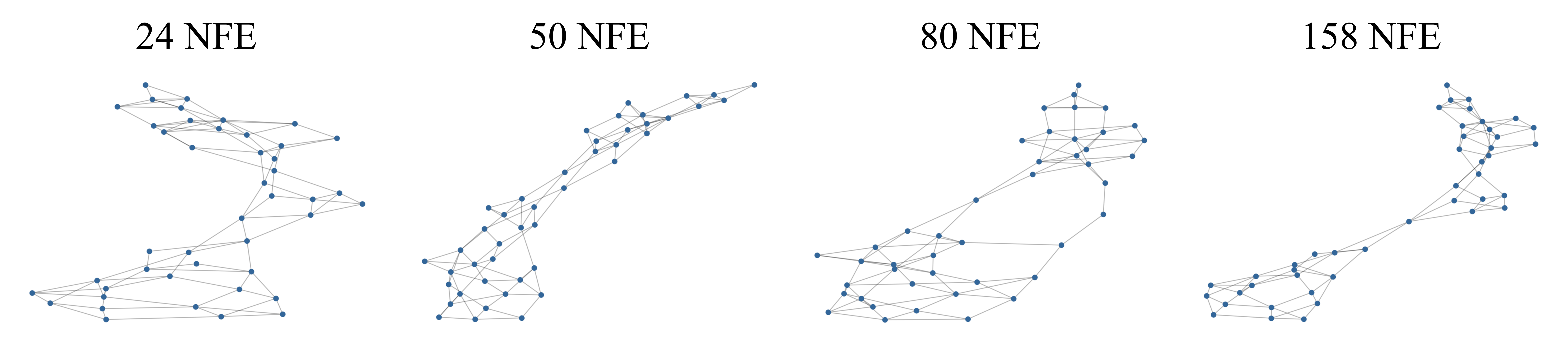}
        \caption{Graphs generated from the same latent code through ODE solvers with various NFE.}
        \label{subfig:ode-vis}
    \end{subfigure}
    \caption{
    Graph generation quality for the same model using different settings of the ODE solvers. 
    The larger error tolerance or step size reduces the number of function evaluations (NFE),  maintaining high generation quality.
    }
    \label{fig:efficient}
\end{figure}

\subsection{Graph Generation Quality}
In this part, we benchmark the sample quality of our proposed graph generative model against other competitive models. 
For implementation details of our GraphGDP, the hidden dimensions are selected from $64$, $128$ and $256$ for different datasets. 
We stack $4$ message passing layers with $8$ attention heads for graph score networks and adopt an MLP with $2$ hidden layers for final score estimation. 
All our models are trained with Adam optimizer \cite{kingma2014adam} and a constant learning rate $2e-5$. We also apply the exponential moving average (EMA) with the momentum $0.9999$ for the parameter updating to improve stability as in \cite{song2021score}.
Notably, the model parameters are shared across time, which is specified to the graph score network using the sinusoidal position embedding \cite{vaswani2017attention}.
At the start of generation, we first sample the number of nodes based on the probability mass function of nodes in the training set.
For datasets with high variance on graph sizes, we could also add the node number information to models as the condition like time information.
We configure the variance preserve SDE with $\bar\beta_{min} = 0.1$ and pick $\bar\beta_{max}$ from $\{5, 10, 20\}$, ensuring that the signal-to-noise ratio at the last perturbed graphs is kept small.
On Ego-small dataset, we apply Euler-Maruyama method with $1000$ discretization steps for graph sampling, and we incorporate extra Langevin MCMC steps on other datasets.  

The generation quality evaluation results with classical structure metrics are reported in Table \ref{tab:main}, where the smaller values of MMD metrics represent the smaller distance between the two distributions.
Table \ref{tab:nn-based} shows the performance using neural-network-based metrics.
We colour the top two performance for each metric.
Some visualizations of generated diffusion processes are shown in Figure \ref{fig:graph_vis}.

We summarize the observations after analyzing the model performance for graph sample quality.
(1) Among order-independent graph generative models, GraphGDP achieves remarkable performance improvement.
(2) Under the same requirement to capture the permutation invariance property of graphs, the proposed GraphGDP outperforms the score-based EDP-GNN obviously, especially in larger graphs. 
(3) Compared to the dominant autoregressive generative models, GraphGDP surpasses the performance of competitive GraphRNN and GRAN for most metrics.
Our method also shows better or comparable results to BIGG without using any predefined node orderings, except for slightly less diversity on the Ego dataset.
In conclusion, GraphGDP demonstrates the high fidelity and diversity of generated graph samples on datasets with different characteristics. 

\subsection{Efficient Graph Generation} \label{subsec:tradeoff}
Sampling efficiency is a crucial property pursued by graph generative models \cite{nips/LiaoLSWHDUZ19, dai2020scalable}. 
For generative diffusion processes, the connection between the reverse-time SDE and the deterministic probability flow ODE provides a way for fast sampling. 
We can generate graphs by solving the neural ODE described by Eq. \ref{eq:ode}.
Through controlling the error tolerance of off-the-shelf adaptive-step solvers or the step size of fixed-step solvers \cite{chen2018neuralode},
we generate high-quality graphs with many fewer steps (a.k.a., function evaluations). 
Compared to other domains, the elements in graph adjacency matrices are less informative with only $2$ values, and may be more tolerant to errors, while the pixels in images have $256$ values.
As shown in Figure \ref{subfig:ode-res}, with different solver settings, our model keeps the structure fidelity with even \textbf{24} steps.
The graph visualization in Figure \ref{subfig:ode-vis} also shows that the overall structure patterns of the graph are maintained even if the graph topology changes slightly due to the numerical precision. 

\begin{figure}[!htbp]
    \centering
    \includegraphics[width=\columnwidth]{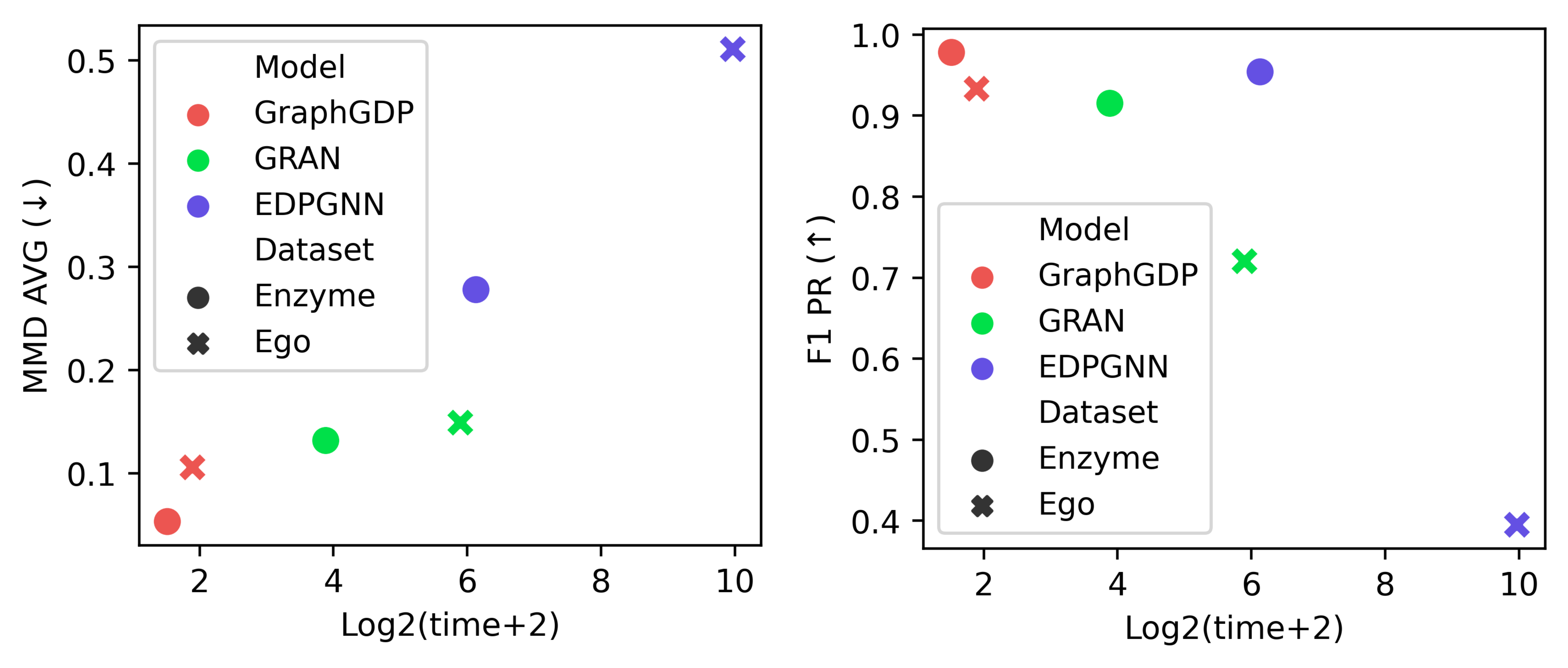}
    \caption{Comparison of the log-scale running time for generating $16$ graphs in a batch on Enzymes and Ego datasets.}
    \label{fig:logtime}
\end{figure}

We compare the sampling quality and inference time of our efficient version with other models in Fig. \ref{fig:logtime}.
GraphGDP has clear performance and speed advantages compared to GRAN (designed for efficient sampling) and EDP-GNN.
Another powerful model, BIGG, does not support graph generation in batch form, and is not put into Fig. \ref{fig:logtime} for comparison.
On Ego dataset, our model takes on average $0.41$s to generate a graph with one batch size, which is still faster than BIGG's $2.19$s.
In contrast to autoregressive models, graph generative diffusion processes have strong potential for efficient generation.

\subsection{Ablation Study}

Utilizing generative diffusion processes, we compare PGSN with its variants and other graph score network parameterization methods on Enzymes dataset to validate the effectiveness of the design for graph score networks. 
All the models are trained with the unified $64$ hidden dimensions and $1$M training steps.
The methods are denoted as follows:
\begin{itemize}
    \item MLP: an MLP consisting of $3$ hidden layers with perturbed graph adjacency matrices as input.
    \item GIN: a $4$-layer graph isomorphism network \cite{xu2018powerful} that takes the degree onehot embedding as the initial node feature. The edge representations are obtained by adding the pair of node representations from the GIN. After concatenating the edge representations and original perturbed edge values, a $2$-layer MLP outputs the estimated scores. 
    \item GAT: a $4$-layer graph attention network \cite{Velickovic2018gat} with $8$ attention heads which follows the same setup of GIN.
    \item GTN: a graph transformer network that modifies the graph attention networks with the dot-product attention and feed-forward network \cite{dwivedi2020generalization}. 
    \item PGSN w/o P: a PGSN variant without node position features consisting of landing probabilities of the node itself, but with the shortest-path-distance features.
    \item PGSN w/o U: a PGSN variant without updating edge features after message passing.
\end{itemize}

From the results in Table \ref{tab:ablation}, it can be observed that using node degree features and the shallow message passing architecture does help the generative models capture the overall degree distribution of the graphs, but fail to go further on the clustering coefficient metric which requires the more accurate graph local structure.  
Notably, introducing the shortest-path-distance features and learnable edge representations greatly improves the ability of the graph score network to denoise the perturbed graphs, while the node position feature from the landing probability of random walks also make contributions.

We also conduct parameter sensitivity analysis for the random walk steps, the results of which are included in Table \ref{tab:ablation}. 
Consistent with the intuition that more random walk steps yield more topology information of graphs, the model with more walks achieves better sample quality.
Considering the computation cost of random walks, we recommend choosing an appropriate step number according to the characteristic of datasets to take into account both efficiency and effectiveness.   

\begin{table}[t]
\caption{The results of various score function parameterization methods in generative diffusion processes on Enzymes dataset.
RW: the steps of random walks in graphs.}
\label{tab:ablation}
\centering
\resizebox{0.9\columnwidth}{!}{%
\renewcommand{\arraystretch}{1.2}
\begin{tabular}{cc|cccc}
\hline
\textbf{Score Network} & \textbf{RW} & \textbf{Deg.} & \textbf{Clus.} & \textbf{Spec.} & \textbf{Avg.} \\ \hline
\hline
MLP     & -  & 0.703          & 1.044          & 0.213          & 0.654          \\
GIN     & -  & 0.331          & 0.468          & 0.071          & 0.290          \\
GAT     & -  & 0.365          & 0.479          & 0.071          & 0.305          \\
GTN      & -  & 0.158          & 0.457          & 0.074          & 0.230          \\ \hline
PGSN w/o P & 32 & 0.083          & 0.213          & 0.050          & 0.116          \\
PGSN w/o U & 32 & 0.107          & 0.212          & 0.048          & 0.122          \\ \hline
PGSN    & 4  & 0.095          & 0.233          & 0.059          & 0.130          \\
PGSN    & 8  & 0.097          & 0.222          & 0.054          & 0.124          \\
PGSN    & 16 & 0.084          & 0.208          & 0.049          & 0.114          \\ 
PGSN    & 32 & \textbf{0.079} & \textbf{0.198} & \textbf{0.047} & \textbf{0.108} \\ \hline
\end{tabular}%
}
\end{table}

\section{Related Work} 
In addition to the graph generation approaches mentioned before, we summarize the notable existing literature on the construction of our framework.

\textbf{Generative Diffusion Processes.}
Diffusion probabilistic models \cite{sohl2015deep} are inspired by non-equilibrium thermodynamics.
By defining a Markov chain of diffusion steps to slowly add noise to data, they learn to reverse the inference path to generate data from the noise. 
Ho \etal \cite{ho2020denoising} propose a simplified objective of diffusion models and connect it with noise-conditioned score networks \cite{song2019generative} which use Gaussian noise to perturb data distribution over the full space.
As it is relatively slow to generate a sample from the Markov chain of the generative diffusion process, a simple stride sampling schedule is proposed by \cite{nichol2021improved}. 
Song \etal \cite{SongDDIM21} define a deterministic generative process and generates high quality samples with a fewer number of steps.
During the same period, Song \etal \cite{song2021score} propose a continuous-time generative diffusion process that takes advantage of the SDE and improves performance.
The flexible model architecture and high sample quality on high-dimensional data attract us to adapt the continuous-time generative process for graph generation.
Most recently, concurrent work \cite{JoLH22GDSS} also studies diffusion models for graph generation but overlooks the discreteness of graphs. 

\textbf{Position-aware Graph Neural Networks.}
Position encoding plays a significant role in current neural networks like ConvNets and Transformers. 
For graph neural networks, position information of nodes or edges is also critical for graph structure representation learning \cite{srinivasan2019equivalence, cui2021positional}.
Adding unique or discriminating features to all nodes in graphs is one way to incorporate the node position \cite{sato2019approximation}. But this type of position encoding lacks the generalization of unseen graphs.  
Laplacian positional encoding takes graph Laplacian eigenvectors that maintain global structure information as external node features \cite{dwivedi2020generalization, kreuzer2021rethinking}. 
The existence of the sign ambiguity is the main weakness of Laplacian positional encoding.
Random walk based position encoding \cite{li2020distance, dwivedi2021lspe} reflects the graph topology by landing probabilities and path distances. 
Inspired by these position-aware GNNs, we realize that the position information in perturbed graphs can reflect the changes in graph structure, and design a position-enhanced graph score network.

\section{Conclusions}
We propose a novel continuous-time generative diffusion process for permutation invariant graph generation.
After constructing a forward diffusion process by an SDE to perturb graph instances towards random graphs, we design a position-enhanced graph score network to extract graph features from hybrid intermediate states, setting up the reverse-time SDE.
We generate high-fidelity and diverse graphs by leveraging the numerical solvers for simulating reverse-time SDE trajectories.
Experiment results show that GraphGDP can generate high-quality graphs in 24 function evaluations for efficient sampling, much faster than autoregressive models.
In the future, we would like to explore this framework with lower computational complexity and further improve efficient sampling. 

\section*{Acknowledgment}
This work was supported in part by the National Natural Science Foundation of China (51991395 and 62272023).

\bibliographystyle{IEEEtran}
\bibliography{ref}

\end{document}